\documentclass[11pt]{ceurart}
\usepackage[T1]{fontenc}
\usepackage{graphics}
\usepackage{float}
\usepackage{hyperref}

\begin{document}

\copyrightyear{2020}
\copyrightclause{Copyright for this paper by its authors.
  Use permitted under Creative Commons License Attribution 4.0
  International (CC BY 4.0).}

\conference{FIRE2020: Forum for Information Retrieval Evaluation, December 16-20, 2020, Hyderabad, India}

\title{JUNLP@Dravidian-CodeMix-FIRE2020: Sentiment Classification of Code-Mixed Tweets using Bi-Directional RNN and Language Tags}

\author{Sainik Kumar Mahata}[
email=sainik.mahata@gmail.com,
]

\author{Dipankar Das}[
email=dipankar.dipnil2005@gmail.com,
]

\author{Sivaji Bandyopadhyay}[
email=sivaji.cse.ju@gmail.com,
]
\address{Computer Science and Engineering, Jadavpur University, Kolkata, India}

\begin{abstract}
Sentiment analysis has been an active area of research in the past two decades and recently, with the advent of social media, there has been an increasing demand for sentiment analysis on social media texts. Since the social media texts are not in one language and are largely code-mixed in nature, the traditional sentiment classification models fail to produce acceptable results. This paper tries to solve this very research problem and uses bi-directional LSTMs along with language tagging, to facilitate sentiment tagging of code-mixed Tamil texts that have been extracted from social media. The presented algorithm, when evaluated on the test data, garnered precision, recall, and F1 scores of 0.59,  0.66, and 0.58 respectively.  
\end{abstract}

\begin{keywords}
  Sentiment Analysis \sep
  LSTM \sep
  Language Tagging
\end{keywords}

\maketitle

\section{Introduction}
Sentiment analysis is the interpretation and classification of emotions (positive, negative, and neutral) within text data using text analysis techniques. It is one of the most important research areas in the domain of Natural Language Processing (NLP) and has garnered much attention in the recent past. However, with the advent of social media, research has become even more wide-spread \cite{10.1007/978-981-13-7403-6_14,garain2020junlpsemeval2020} as it takes into account conversations of customers around the social space and puts them into context. But, in the context of the Indian subcontinent, much research has been focused on sentiment classification of social media texts that are generally code-mixed in nature. This is because India has a linguistically diverse diaspora and a large number of the Indian population is comfortable in more than one language \cite{10.1145/3368567.3368579}. This leads to communication in sentences, which contain more than one language in the same phrase \cite{MANDAL18.27}. Furthermore, words of different languages are generally written in Roman Script, which leads to the formation of a complex syntax structure which is difficult to parse with traditional NLP tools.

This paper aims to solve this research problem and uses Bi-Directional LSTMs \cite{10.1162/neco.1997.9.8.1735} to tag the texts with its respective sentiment. Language tagging of individual words was used as additional features while training our classification model. Moreover, the training corpus was passed through FastText \cite{bojanowski2016enriching} embedding, to map the semantically similar words in a common 3D space. The designed model was evaluated on the test data and garnered an F1 score of 0.58. The code of our developed model is available as a git repository  \href{https://github.com/sainikmahata/FIRE-Dravidian-sentiment_analysis.git}{here}.

This model was designed as a part of the "Dravidian-CodeMix - FIRE 2020\footnote{https://dravidian-codemix.github.io/2020/}" shared task and was evaluated for English-Tamil code-mixed texts. The goal of this task was to identify sentiment polarity of the code-mixed dataset of comments/posts in Dravidian Languages (Malayalam-English and Tamil-English) collected from social media. 

The rest of the paper is organized as follows. Section \ref{sec:Data} deals with the description of the training data that was used to build the proposed sentiment classification system. Section \ref{sec:Framework} describes the proposed model and the various sub-models used to build the system and will be followed by the evaluation of the model in Section \ref{sec:Evaluation}. \nocite{*}

\section{Data}
\label{sec:Data}
The organizers provided us with Tamil-English and Malayalam-English code-mixed text data, derived from YouTube video comments. The dataset contained all the three types of code-mixed sentences -- Inter-Sentential switch, Intra-Sentential switch, and Tag switching and had 5 output labels; Positive, Negative, Mixed Feelings, Not Tamil, and Unknown State. Most comments were written in Roman script with either Tamil / Malayalam grammar with English lexicon or English grammar with Tamil / Malayalam lexicon. Some comments were written in Tamil / Malayalam script with English expressions in between. We participated in the sentiment classification of the code-mixed English-Tamil text task only. The English-Tamil dataset was divided into training, validation and test data which had 11,335, 1,260 and 3,149 code-mixed sentence instances respectively..

\section{Framework}
\label{sec:Framework}
Initially, the training and validation data,  without the sentiment labels, were merged together, tokenized using the NLTK\footnote{https://www.nltk.org/} library, and this was used to extract word vectors of size 100, using the FastText\footnote{https://fasttext.cc/} embedding. The skip-gram model was used instead of the continuous-bag-of-words (CBOW) model as skip-gram works best for low data sizes. The model took into account character n-grams from 3 to 6 characters. 

After this step, the data was preprocessed to find out the language tags of individual words in a sentence. For this, the nltk corpus was used. The words were given as input and were tagged as English or non-English by the algorithm. 

Thereafter, vectors of sentences of the train and validation dataset were extracted from the trained embedding. The language tags and the words vectors were merged together using a Concatenation layer and were given as input to a Bi-Directional LSTM cell. The context vector was then mapped to the output labels with the help of a Dense layer.

\begin{figure}
    \centering
    \includegraphics[scale=0.7]{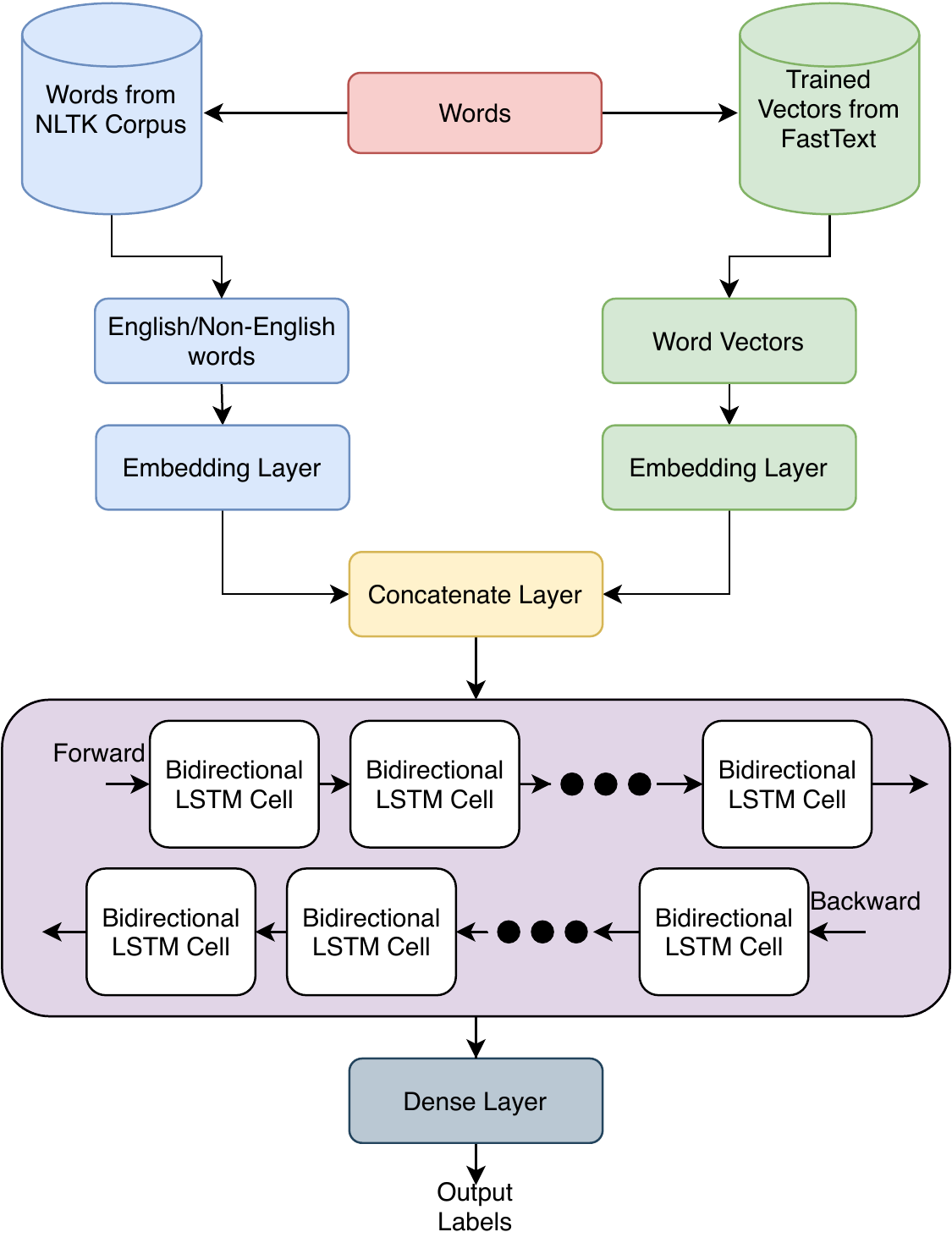}
    \caption{Code-Mixed Sentiment Analysis model.}
    \label{fig:my_label}
\end{figure}

The schematic of the model is shown in Figure \ref{fig:my_label}. Other parameters of the model are as follows.
\begin{itemize}
    \item batch size: 32
    \item epochs: 50
    \item optimizer: adam
    \item loss: sparse categorical cross-entropy
    \item validation split: 0.1
\end{itemize}

\section{Evaluation}
\label{sec:Evaluation}
On testing the model on the initial test dataset, the model garnered accuracy and F1-Score of 70.42\% and 0.63 respectively. We also trained three other models, where the basic architecture was the same, the difference being the usage of LSTM/Bi-Directional LSTM and language tag features. The models were 
\begin{itemize}
    \item Bidirectional LSTM without the language tag feature.
    \item LSTM with the language tag feature.
    \item LSTM without the language tag feature.
\end{itemize} 
The accuracy and F1-Score of every model are shown in Table 1.

\begin{table}
\label{tab:comparison}
\caption{Comparison of accuracy scores of all the developed models.}
\begin{tabular}{ccccc}
\hline
\textbf{Model}     & \textbf{Bi-LSTM+ln tag} & \textbf{Bi-LSTM} & \textbf{LSTM+ln tag} & \textbf{LSTM} \\ \hline
\textbf{Accuracy}  & 70.42\%                 & 70.82\%          & 70.62\%              & 70.22\%       \\
\textbf{F1-Score}  & 0.63                    & 0.61             & 0.62                 & 0.62          \\
\textbf{Precision} & 0.62                    & 0.59             & 0.63                 & 0.62          \\
\textbf{Recall}    & 0.70                    & 0.71             & 0.71                 & 0.70          \\ \hline
\end{tabular}
\end{table}

We chose the Bi-Directional LSTM model with language tag features as the final model, as it has the best F1-Score out of all the developed models.

The system was submitted to the organizers and was evaluated using the gold standard dataset, developed by the organizers. The results of the evaluation are shown in Table 2.

\begin{table}
\label{tab:organizer_eval}
\caption{Final evaluation result by the Organizers.}
\begin{tabular}{cccc}
\hline
\textbf{Team Name} & \textbf{Precision} & \textbf{Recall} & \textbf{F1-Score} \\ \hline
JUNLP              & 0.59               & 0.66            & 0.58              \\ \hline
\end{tabular}
\end{table}

\section{Conclusion}
In the current work, we attempted to solve the problem of Sentiment Analysis of code-mixed English-Tamil data, while participating in the Dravidian-Code-Mixed-FIRE2020 shared task. Our system was based on using Bi-Directional LSTM along with Language Tag features. Also, FastText embedding was used to generate word vectors to train the model. Our system, when evaluated by the organizers garnered an F1 score of 0.58. As future work, we would like to increase this data, use state-of-the-art Neural Network architectures, like BERT, RoBERTa, etc., on this data, taking into advantage the concept of matrix and embedded language, SentiWordNet, and other NLP features.

\section{Acknowledgment}
This work is supported by Digital India Corporation, MeitY, Government of India, under the Visvesvaraya PhD Scheme for Electronics \& IT.

\bibliography{sample-ceur}

\end{document}